\documentclass[runningheads]{llncs}

\usepackage{eccv}

\usepackage{eccvabbrv}

\usepackage{graphicx}
\usepackage{booktabs}
\usepackage[belowskip=-7pt,aboveskip=3pt]{caption}
\usepackage{multirow}
\usepackage{tabularx}
\newif\ifdrafting
\draftingtrue % To show comments
\ifdrafting

\else

\fi

\definecolor{guppiegreen}{rgb}{0.0, 0.7, 0.7}
\definecolor{alizarin}{rgb}{0.8, 0.1, 0.2}
\definecolor{brandeisblue}{rgb}{0.0, 0.44, 1.0}

\newcommand{\st}[1]{\textbf{#1}}
\newcommand{\nd}[1]{\underline{#1}}

\usepackage[pagebackref,breaklinks,colorlinks,citecolor=eccvblue]{hyperref}
\usepackage{microtype}  % character protrusion + font expansion for flush paragraphs

\makeatletter
\@ifpackageloaded{eccvabbrv}{}{
  \newcommand{\eg}{\emph{e.g.}\xspace}
  \newcommand{\ie}{\emph{i.e.}\xspace}
  \newcommand{\etal}{\emph{et al.}\xspace}

}
\makeatother

\usepackage[capitalize,noabbrev]{cleveref}

\AtBeginDocument{%
  \crefname{figure}{Fig.}{Figs.}%
  \Crefname{figure}{Fig.}{Figs.}%
  \crefname{table}{Table}{Tables}%
  \Crefname{table}{Table}{Tables}%
  \crefname{section}{Section}{Sections}%
  \Crefname{section}{Section}{Sections}%
  \crefformat{equation}{Eq.~(#2#1#3)}%
  \Crefformat{equation}{Eq.~(#2#1#3)}%
  \crefrangeformat{equation}{Eqs.~(#3#1#4) to~(#5#2#6)}%
  \crefmultiformat{equation}{Eqs.~(#2#1#3)}{ and~(#2#1#3)}{, (#2#1#3)}{ and~(#2#1#3)}%
  \Crefmultiformat{equation}{Eqs.~(#2#1#3)}{ and~(#2#1#3)}{, (#2#1#3)}{ and~(#2#1#3)}%
}

\begin{document}

\title{Fidelity- and Perception-Aware \\ Local Implicit Attention for \\ Arbitrary-Scale Image Super-Resolution}

\titlerunning{Fidelity- and Perception-Aware Local Implicit Attention for ASISR}

\author{
Yu-Syuan Xu\inst{1, 2} \ Hao-Lun Sun\inst{2} \  Hao-Wei Chen\inst{1} \\ 
Hsien-Kai Kuo\inst{2} \  Chun-Yi Lee\inst{1}
\\
{\tt\small \{Yu-Syuan.Xu, Hsienkai.Kuo\}@mediatek.com} \\
{\tt\small \{d13922023,cylee\}@csie.ntu.edu.tw} \\
}

\authorrunning{Yu-Syuan Xu \and  Hao-Lun Sun \and  Hao-Wei Chen \and  Hsien-Kai Kuo \and  Chun-Yi Lee}

\institute{
\textsuperscript{1}National Taiwan University \quad 
\textsuperscript{2}MediaTek Inc.
}

\maketitle

\begin{abstract}
% \vspace{-20pt}
Arbitrary-scale image super-resolution (ASISR) aims to reconstruct high-resolution images from low-resolution inputs over a continuous range of upscaling factors. 
While traditional pixel-regression approaches often produce overly smooth results that lack realistic details, recent diffusion  methods can produce sharper and more realistic textures. However, these diffusion techniques frequently introduce the risk of structural hallucinations. To address these issues, we propose Fidelity- and Perception-Aware Local Implicit Attention (FPLIA), a framework that effectively integrates fidelity-oriented features into a diffusion  pipeline to produce realistic and faithful reconstructions for ASISR. 
We introduce a Fidelity and Perception Attention Module (FPAM), which applies both self-attention and cross-attention to fidelity-oriented and perceptual features to enhance representational capacity. To further exploit their complements, we design a Fidelity and Perception Select Module (FPSM) that adaptively selects the most representative features for RGB values prediction. We conduct extensive experiments to validate the effectiveness of these components. Both qualitative and quantitative results show that FPLIA delivers superior perceptual realism while maintaining reconstruction accuracy on standard ASISR benchmarks. The source code is accessible at the following repository: \href{https://github.com/XUSean0118/FPLIA}{https://github.com/XUSean0118/FPLIA}. 

\vspace{-5pt}
\end{abstract}
% ============================================================
% ORIGINAL STUDENT FIGURE CAPTION (commented out — do not delete)
% ============================================================
% \begin{figure}[t]
%     \centering
%     \includegraphics[width=1\columnwidth]{figures/intro.pdf}
%     \caption{Comparison of regression-based and generation-based approaches with FPLIA. FPLIA adaptively harnesses the strengths of both approaches while avoiding their weaknesses: the over-smoothing typical of regression-based methods (e.g., the blinds in the first row) and the hallucination artifacts of diffusion methods (e.g., spurious 'o' replacing 't' in "create" and 'm' replacing 'n' in "presentation" in the second row).}
%     \label{fig::intro}
%     \vspace{-10pt}
% \end{figure}
% ============================================================

\begin{figure}[t]
    \centering
    \includegraphics[width=1.0\columnwidth]{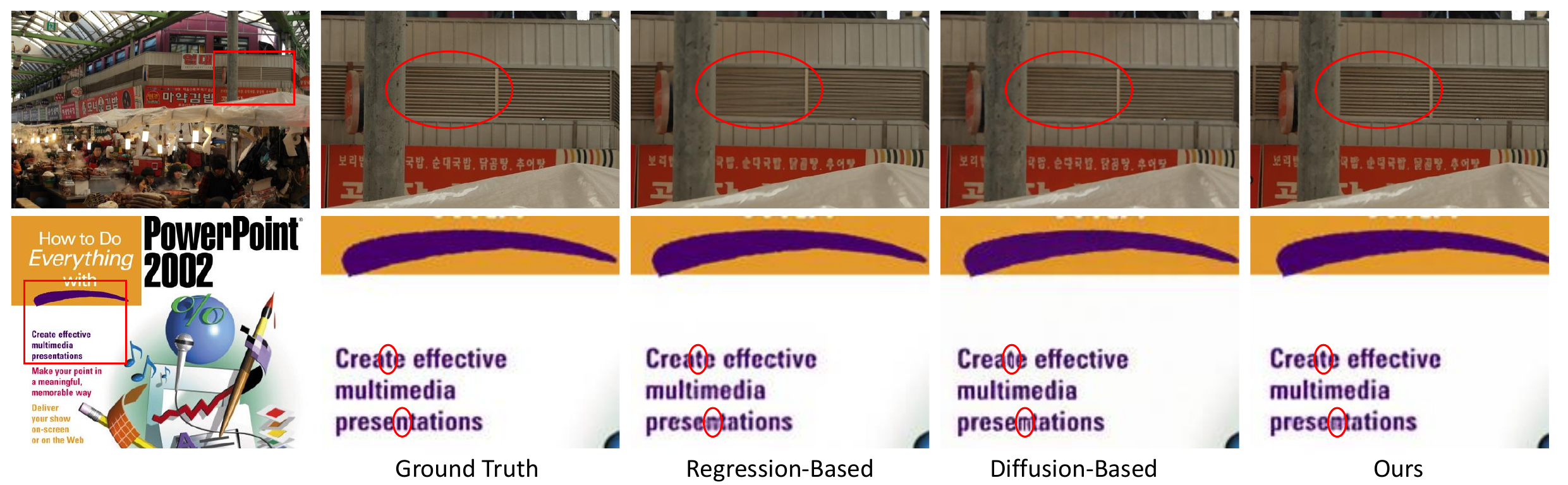}
    \caption{Regression-based and diffusion-based approaches for arbitrary-scale
    image super-resolution (ASISR) exhibit complementary failure modes.
    Regression methods (\eg, LIT~\cite{clit} with SwinIR~\cite{swinir})
    reconstruct structurally faithful images that suppress fine textures,
    such as the blinds in the first row.
    Diffusion methods (\eg, Kim \etal~\cite{kim2024} with LDM~\cite{ldm})
    recover perceptually realistic textures, yet may hallucinate content
    absent from the ground truth, such as the spurious characters in the
    second row. FPLIA harnesses both feature types through cross-feature
    attention and adaptive per-pixel selection, producing reconstructions
    that are both faithful and perceptually sharp.}
    \label{fig:intro}
    \vspace{-10pt}
\end{figure}

\vspace{-10pt}
\section{Introduction}
\label{sec:intro}

Arbitrary-scale image super-resolution (ASISR) aims to reconstruct high-resolution (HR) images from low-resolution (LR) inputs at any continuous upscaling factor within a single model~\cite{metasr, liif, ultrasr, itsrn, ipe, lte, wang2023, clit, ciaosr, hiif, linf, idm, kim2024}. 
This formulation provides a practical and flexible alternative to fixed-scale super-resolution methods~\cite{srcnn, edsr, rdn, rcan, swinir, esrgan, sr3}, which require training and maintaining a separate network for each target resolution. 
A prevailing paradigm in ASISR combines deep feature extractors with local implicit image functions, \ie, coordinate-conditioned networks that map 2D query positions and their
associated latent features to RGB pixel values~\cite{liif, lte, clit, ciaosr, hiif, idm, kim2024}. 
The reconstruction quality of such methods depends critically on the nature of the underlying latent representations. 
As \cref{fig:intro} illustrates, features extracted by pixel-regression backbones~\cite{edsr, rdn, swinir} trained with $\ell_1$ or $\ell_2$ objectives yield images with high pixel-wise fidelity, yet the reconstructions tend to suppress fine-grained textures and high-frequency details. 
Features generated by diffusion models~\cite{sr3, ldm}, by contrast, recover perceptually convincing textures, yet they risk introducing hallucinated structures that deviate from the ground-truth content. 
This fundamental tension between reconstruction fidelity and perceptual realism poses two intertwined challenges for ASISR. 
First, fidelity-oriented and perceptual features encode substantially different information: one prioritizes low-level pixel accuracy, while the other captures distributional statistics that favor visual realism. 
Facilitating productive interaction between these heterogeneous representations without mutual degradation remains an open problem. 
Second, the relative utility of each feature type is not spatially uniform: structurally simple regions benefit primarily from fidelity features, whereas richly textured areas demand stronger perceptual cues. 
A framework that fuses the two in a content-agnostic manner inevitably compromises one objective for the other across different image regions.

Existing ASISR approaches have addressed the fidelity-perception divide primarily by committing to one paradigm.
Regression-based methods such as LIIF~\cite{liif}, LTE~\cite{lte}, CiaoSR~\cite{ciaosr},
LIT~\cite{clit}, and HIIF~\cite{hiif} pair local implicit functions with pixel-regression backbones and achieve strong fidelity-oriented performance, yet consistently underperform on perceptual metrics, as the regression objective inherently suppresses high-frequency variations.
Diffusion-based approaches such as IDM~\cite{idm} and Kim \etal~\cite{kim2024} replace the regression backbone with a generative model, improving perceptual quality at the expense of reconstruction fidelity; hallucinated textures and structural distortions arise when the generative process lacks sufficient structural constraint from the input, particularly in regions with ambiguous or repetitive content.
Regarding cross-feature interaction, Kim \etal~\cite{kim2024}, the closest effort toward bridging the two paradigms, conditions its diffusion pipeline on regression-derived features, yet the two streams remain loosely coupled, without mechanisms for bidirectional information exchange.
The ASISR landscape thus remains divided along the fidelity-perception axis.

Beyond the surface-level fidelity-perception dichotomy identified above, the complementarity between the two feature types exhibits a finer structure governed by two axes of variation.
Along the spatial axis, the relative informativeness of each feature type depends on local image content: structurally simple regions are already well-served by fidelity features, while richly textured areas require perceptual cues to recover plausible high-frequency details.
Along the scale axis, fidelity features grow progressively less informative at higher upscaling factors, as the LR input retains less high-frequency content and perceptual features become increasingly necessary to synthesize fine details that the fidelity stream can no longer provide.
This dual variability implies that no fixed combination strategy can serve all spatial locations and scale factors simultaneously.
Effective integration instead requires two capabilities: cross-feature interaction that produces diverse candidate representations spanning different fidelity-perception profiles, and per-pixel selection that identifies the most reliable candidate at each queried coordinate.
The interaction between the two feature streams is itself asymmetric in nature.
Fidelity features provide a structurally anchored scaffold capable of guiding perceptual features toward more faithful textures, while perceptual features inject high-frequency diversity that enriches otherwise smooth fidelity representations.
This asymmetry suggests that the two directions of information exchange between the streams should be modeled as separate pathways rather than through a single symmetric operation such as feature concatenation.

Building on these observations, this paper presents \textbf{Fidelity- and Perception-aware Local Implicit Attention (FPLIA)}, a framework that integrates fidelity-oriented and perceptual features for ASISR through explicit cross-feature interaction and adaptive per-pixel feature selection.
FPLIA addresses the two challenges above with two complementary modules.
The Fidelity and Perception Attention Module (FPAM) models asymmetric bidirectional interaction between the two feature streams, producing four candidate representations at every queried coordinate, each capturing a different balance between pixel-level accuracy and perceptual realism.
The Fidelity and Perception Select Module (FPSM) estimates the reliability of each candidate by combining content-dependent attention cues with scale-dependent signals, and commits to the single highest-confidence candidate at every position.
% This combination allows FPLIA to adapt its feature usage to both local content and scale factor without manual tuning or fixed blending ratios.
The framework is designed to be backbone-agnostic: the fidelity-oriented extractor and the perceptual generator can be instantiated with different architectures, as demonstrated with several representative choices in the supplementary material.
Extensive experiments across integer and non-integer upscaling factors on DIV2K~\cite{div2k}, Set5~\cite{set5}, Set14~\cite{set14}, B100~\cite{b100}, and Urban100~\cite{urban100} confirm that FPLIA achieves a Pareto-optimal balance between fidelity and perceptual quality, simultaneously improving both PSNR and LPIPS over the leading diffusion-based methods across the majority of settings. The key contributions of this work are as follows:
\vspace{-5pt}
\begin{itemize}
\item Interactions between fidelity and perceptual feature streams produce more diverse latent representations than either stream in isolation.
This finding motivates the design of FPAM, which employs self-attention within each branch and cross-attention across branches to generate four candidate features with distinct fidelity-perception profiles at every queried coordinate.

\item Adaptive per-pixel feature selection, guided by confidence scores derived from attention maps and scale-dependent cues, enables content-aware reconstruction.
The proposed FPSM estimates the reliability of each candidate at every queried coordinate, allowing the framework to favor fidelity features in structurally simple regions and perceptual features in texture-rich areas.

\item FPLIA achieves Pareto-optimal fidelity-perception balance on standard ASISR benchmarks, improving LPIPS by up to $29.6\%$ over the best regression-based methods while maintaining a PSNR trade-off within $2.6\%$, and simultaneously improving both metrics over the leading diffusion-based methods, with less than $5\%$ latency and $2\%$ memory overhead.
\end{itemize}

% \noindent Related work is provided in the supplementary material. \Cref{sec:preliminary} introduces background, \cref{sec:method} presents the proposed framework, \cref{sec:experiment} reports experimental results, and \cref{sec:conclusion} concludes.

\section{Preliminary}
\label{sec:preliminary}

\subsection{Local Implicit Image Functions}
\label{subsec:liif_background}

ASISR seeks to reconstruct a high-resolution image from a low-resolution input at any upsampling factor within a single model. 
A widely adopted paradigm for this task represents images as continuous functions defined over a 2D coordinate space $\mathcal{X}$~\cite{liif, lte, clit, ciaosr, hiif}. 
Given a low-resolution image, a deep feature extractor first produces a feature map. 
For each queried high-resolution coordinate $x_q \in \mathcal{X}$, the framework identifies the nearest feature positions on the low-resolution grid and retrieves the corresponding latent vectors. 
% These local features are concatenated with the relative coordinate offset $\delta x$ between $x_q$ and each grid position,
These local features are concatenated with the cell size $c$ that encodes the spatial extent of a pixel at the target resolution. 
A coordinate-conditioned decoder, typically implemented as a multi-layer perceptron, then maps this concatenated representation to an RGB value at $x_q$. 
Because the decoder operates on continuous coordinates rather than fixed pixel locations, a single trained model can serve arbitrary upsampling factors. 
% The reconstruction quality of this paradigm, however, depends critically on the expressiveness of the underlying feature map, and existing methods derive their features exclusively from a single pixel-regression backbone trained with $\ell_1$ or $\ell_2$ objectives~\cite{edsr, rdn, swinir}, leaving the decoded output upper-bounded by what this single feature stream can represent.

\subsection{Cross-Scale Local Attention}
\label{subsec:csla_background}

To move beyond uniform feature aggregation, recent extensions of the local implicit paradigm replace the simple concatenation scheme described above with an attention-based mechanism that computes content-dependent weightings over the local neighborhood of each queried coordinate~\cite{clit, ciaosr}. 
The feature map is first projected into query, key, and value embeddings through convolutional layers. 
The query embedding $q \in \mathbb{R}^{1 \times C}$ at a target coordinate $x_q$ is obtained via bilinear interpolation, while key and value embeddings $k, v \in \mathbb{R}^{G_h G_w \times C}$ are sampled over a local grid $x^{G_h \times G_w}$ of size $G_h \times G_w$ centered at $x_q$. 
To encode the geometric relationship between the query and each grid position, the relative offset $\delta x_{ij} = x_q - x_{ij}$ is passed through a positional encoding function $\gamma$ followed by a fully connected layer to yield a positional bias $B = \mathrm{FC}(\gamma(\delta x))$. 
An attention map $a = \mathrm{Softmax}(q k^\top / \sqrt{C} + B)$ is then formed, and the attended feature is computed as $\tilde{f} = \mathrm{FC}(\mathrm{Concat}(a \times v,\; c))$, where the cell size $c$ is appended before the final projection so that the aggregation adapts to the target resolution. 
This mechanism allows the decoder to focus on the most informative neighbors for each query and to modulate its response according to the upscaling factor.
% The existing formulation, however, applies this attention to a single feature stream; when multiple feature sources with distinct statistical characteristics are available, attending over them within the same operation conflates representations from different manifolds, and a revised design is needed to accommodate such heterogeneity.

\section{Methodology}
\label{sec:method}

% ============================================================
% ORIGINAL STUDENT §3 opening (commented out — do not delete)
% In this section, we first present an overview of the Fidelity and Perception Local Implicit Attention (FPLIA) framework in \cref{subsec::overview}. Subsequently, we elaborate on our proposed Fidelity and Perception Attention Module (FPAM) and Fidelity and Perception Select Module (FPSM) in \cref{subsec::fpam,subsec::fpsm}, respectively.
% ============================================================

% --- §3 opening ---
% The remainder of this section develops the proposed framework in four stages. 
% \Cref{subsec:problem} formulates the reconstruction problem and identifies the central design challenge. 
% \Cref{subsec:analysis} analyzes the structural properties of fidelity-oriented and perceptual features that inform the architectural choices. 
% \Cref{subsec:overview} provides am overview of the complete pipeline, followed by detailed descriptions of the Fidelity and Perception Attention Module in \Cref{subsec:fpam} and the Fidelity and Perception Select Module in \Cref{subsec:fpsm}. 
% \Cref{subsec:training} presents the training objectives.

% --- §3.1 Problem Formulation ---
\subsection{Problem Formulation}
\label{subsec:problem}

Given a low-resolution image $I^{LR} \in \mathbb{R}^{H \times W \times 3}$, the goal is to reconstruct a high-resolution image $I^{HR} \in \mathbb{R}^{r_h H \times r_w W \times 3}$ under an arbitrary upsampling factor $r = \{r_h, r_w\}$. 
Following the local implicit paradigm reviewed in \Cref{subsec:liif_background}, reconstruction proceeds by predicting the RGB value at each queried coordinate $x_q$ in the continuous space $\mathcal{X}$. 
The framework has access to two complementary feature sources: fidelity-oriented features $\mathcal{F}_f \in \mathbb{R}^{H \times W \times C}$, extracted by a pixel-regression backbone, and perceptual features $\mathcal{F}_p \in \mathbb{R}^{H \times W \times C}$, produced by a diffusion-based generator. %conditioned on $\mathcal{F}_f$. 
The two streams encode fundamentally different information: $\mathcal{F}_f$ preserves pixel-level accuracy at the cost of high-frequency detail, while $\mathcal{F}_p$ recovers perceptually convincing textures at the risk of structural hallucination. 
A direct combination of the two through fixed blending or simple concatenation treats all spatial locations and scale factors uniformly, an assumption that conflicts with the observation that the relative utility of each feature type varies with both local image content and the upscaling factor. 
Addressing this challenge is the central motivation for the architectural choices developed in the following subsections.

% --- §3.2 Design Analysis ---
\subsection{Design Analysis}
\label{subsec:analysis}

As established in \Cref{sec:intro}, the interaction between fidelity-oriented and perceptual features is inherently asymmetric. The concrete consequence for attention-based interaction is directional: when fidelity features attend to perceptual keys, the resulting representation inherits fine-grained detail while remaining spatially grounded; when perceptual features attend to fidelity keys, the result trades some textural richness for structural consistency. These two information flows address different failure modes and should therefore be modeled as separate, directional pathways rather than collapsed into a single symmetric operation.

\begin{remark}[Asymmetric cross-feature interaction]
Let $\tilde{f}_{fp}$ denote the feature obtained when a fidelity-derived query attends to perceptual keys and values, and let $\tilde{f}_{pf}$ denote the converse. The two cross-attention outputs occupy distinct regions of the fidelity-perception spectrum: $\tilde{f}_{fp}$ supplements a fidelity-anchored representation with perceptual detail, whereas $\tilde{f}_{pf}$ constrains a perception-driven representation with structural fidelity. Neither subsumes the other, and both complement the self-attention-refined features $\tilde{f}_f$ and $\tilde{f}_p$, which remain within their respective original manifolds.
\end{remark}

The four candidate features produced by this architecture each encode a different fidelity-perception balance, and their diversity is valuable only if the framework can commit to the most suitable one at each coordinate rather than averaging across them. Soft blending is insufficient for this purpose: features from different statistical manifolds can produce degenerate outputs when averaged, and a globally learned weighting forfeits the spatial adaptivity the architecture is designed to provide. The optimal candidate varies with both local content and upscaling factor, as confirmed by the analysis in \Cref{subsec::selection_analysis}. A discrete, per-pixel selection conditioned on content-derived attention cues and scale-encoding cell size $c$ is therefore the natural design choice.

\begin{figure}[t]
    \centering
    \includegraphics[width=0.9\linewidth]{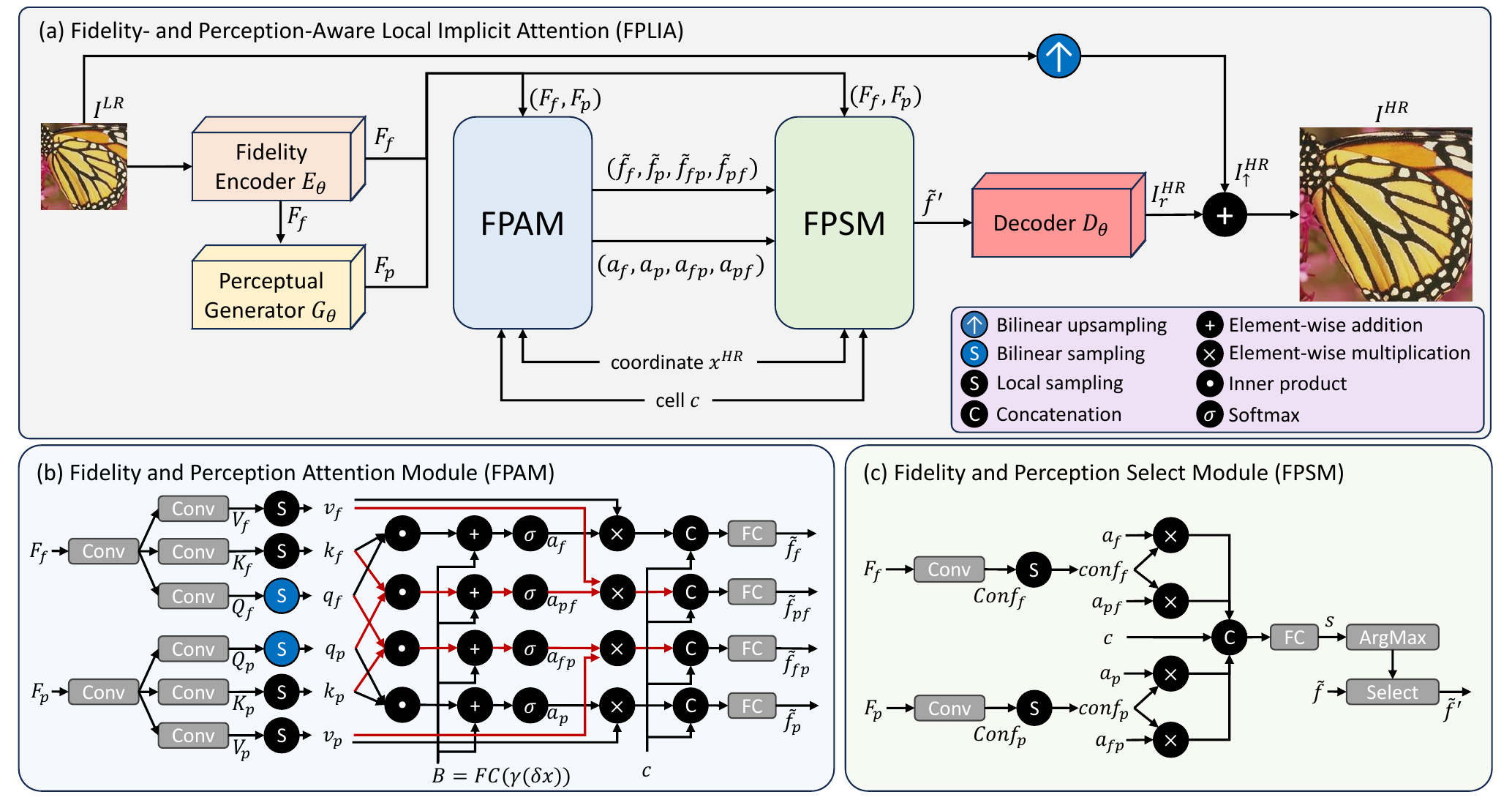}
    \caption{Overview of the proposed FPLIA framework. (a) FPLIA jointly exploits fidelity-oriented and perceptual features for advancing ASISR. (b) FPAM employs self-attention and bi-directional cross-attention mechanisms to produce enhanced representative latent features. (c) FPSM estimates confidence scores and adaptively selects higher-confidence features for predicting RGB values at each queried coordinate.}
    \label{fig:framework}
    \vspace{-10pt}
\end{figure}

\vspace{-5pt}
\subsection{Framework Overview}
\label{subsec:overview}

Guided by the analysis above, FPLIA instantiates these principles in a five-component pipeline illustrated in \Cref{fig:framework}: a fidelity-oriented feature extractor $E_\theta$, a perceptual feature generator $G_\theta$ conditioned on $\mathcal{F}_f$, FPAM, FPSM, and a decoder $D_\theta$ that maps the selected feature to a residual RGB prediction added to a bilinearly upsampled $I^{LR}$. Both $E_\theta$ and $G_\theta$ are modular plug-ins; \Cref{subsec:fpam,subsec:fpsm} detail FPAM and FPSM respectively.

\vspace{-5pt}
\subsection{Fidelity and Perception Attention Module}
\label{subsec:fpam}

To realize the asymmetric bidirectional interaction established in Remark~1, FPAM extends the single-stream cross-scale local attention reviewed in \Cref{subsec:csla_background} to a dual-stream architecture that operates jointly over $\mathcal{F}_f$ and $\mathcal{F}_p$. Both feature maps are projected through convolutional layers into query, key, and value embeddings $(Q_f, K_f, V_f)$ and $(Q_p, K_p, V_p)$. For a queried coordinate $x_q \in x^{HR}$, the query embeddings $q_f, q_p \in \mathbb{R}^{1 \times C}$ are bilinearly interpolated from $Q_f$ and $Q_p$ at $x_q$, while the local key and value embeddings $(k_f, v_f)$ and $(k_p, v_p) \in \mathbb{R}^{G_h G_w \times C}$ are sampled from their respective maps over a local grid $x^{G_h \times G_w}$ centered at $x_q$. The relative positional offset $\delta x_{ij} = x_q - x_{ij}$ for each grid position $x_{ij}$ and the cell size $c = (2/r_h, 2/r_w)$ are shared across all four attention pathways. Applying the $\mathrm{CSLA}(\cdot)$ function defined in \Cref{subsec:csla_background} in four configurations yields:
\begin{equation}
    \begin{alignedat}{2}
        \tilde{f}_f,\; a_f &= \mathrm{CSLA}(q_f, k_f, v_f, \delta x, c), &\qquad
        \tilde{f}_p,\; a_p &= \mathrm{CSLA}(q_p, k_p, v_p, \delta x, c), \\
        \tilde{f}_{pf},\; a_{pf} &= \mathrm{CSLA}(q_p, k_f, v_f, \delta x, c), &\qquad
        \tilde{f}_{fp},\; a_{fp} &= \mathrm{CSLA}(q_f, k_p, v_p, \delta x, c),
    \end{alignedat}
\label{eq:fpam_branches}
\end{equation}
where $\tilde{f}_f, \tilde{f}_p \in \mathbb{R}^{1 \times C}$ are the self-attention outputs that refine each stream within its own statistical manifold, and $\tilde{f}_{pf}, \tilde{f}_{fp} \in \mathbb{R}^{1 \times C}$ are the cross-attention outputs that produce the asymmetric hybrid features described in Remark~1. The corresponding attention maps $a_{(\cdot)} \in \mathbb{R}^{1 \times G_h G_w}$ are retained for use in the downstream selection stage.

\subsection{Fidelity and Perception Select Module}
\label{subsec:fpsm}

Following the content- and scale-dependent selection rationale developed in \Cref{subsec:analysis}, FPSM identifies the most reliable candidate feature at each queried coordinate $x_q$ by estimating a confidence score for each of the four outputs produced by FPAM. To capture the reliability of each feature stream at a global spatial level, FPSM first predicts dense confidence maps $Conf_f$ and $Conf_p$ from the enhanced feature maps $\mathcal{F}_f$ and $\mathcal{F}_p$ through convolutional layers, and samples local confidence values $conf_f$ and $conf_p$ on the grid around $x_q$. These confidence values are combined with the attention maps $a_f, a_p, a_{pf}, a_{fp}$ from FPAM, which encode the local content structure around the query, and with the cell size $c$, which encodes the target scale. The per-candidate confidence scores $s = [s_f, s_p, s_{pf}, s_{fp}]$ are computed as:
\begin{equation}
    s = \mathrm{FC}\!\left(\mathrm{Concat}(a_f \times conf_f,\; a_p \times conf_p,\; a_{pf} \times conf_f,\; a_{fp} \times conf_p,\; c)\right),
\label{eq:confidence_score}
\end{equation}
where the element-wise products $a_{(\cdot)} \times conf_{(\cdot)}$ fuse content-level attention cues with stream-level reliability estimates. Conditioning on both signals enables FPSM to adapt its selection along the spatial and scale axes analyzed in \Cref{subsec:analysis}.

Given the confidence scores $s$, FPSM selects the highest-confidence candidate through an $\arg\max$ operation. Because discrete selection is non-differentiable, the Gumbel-Softmax relaxation~\cite{gumbel} is adopted during training to enable gradient propagation. A soft one-hot vector is formed as $s' = \mathrm{Softmax}((s + \Delta) / \tau)$, where $\Delta \sim \mathrm{Gumbel}(0, 1)$ is stochastic noise and $\tau > 0$ is a temperature that is initialized to $1$ and annealed to $0.001$ during training. The selected feature $\tilde{f}'$ at $x_q$ is then obtained as a weighted sum during training and as a hard selection at inference:
\begin{equation}
    \tilde{f}' = \sum_{i \in \{f,\, p,\, pf,\, fp\}} s'_i \times \tilde{f}_i \quad \text{(training)}, \qquad \tilde{f}' = \tilde{f}_{j^\star},\; j^\star = \arg\max_j\, s_j \quad \text{(inference)}.
\label{eq:selection}
\end{equation}
Temperature annealing gradually sharpens the soft distribution toward this hard regime, ensuring a smooth transition between training and inference behavior.

% --- §3.6 Training Objectives ---
\subsection{Training Objectives}
\label{subsec:training}

The decoder $D_\theta$ is applied not only to the selected feature $\tilde{f}'$ but also independently to each of the four candidate features, producing five predictions at every sampled coordinate: the selected output $I_k^{HR}$, the fidelity output $I_k^{HR_f}$, the perceptual output $I_k^{HR_p}$, the perception-to-fidelity output $I_k^{HR_{pf}}$, and the fidelity-to-perception output $I_k^{HR_{fp}}$. Supervising all five branches ensures that each candidate is trained to be independently informative, which in turn provides FPSM with a meaningful selection landscape in which confidence scores reflect genuine differences in reconstruction quality rather than artifacts of unbalanced training. For each sampled pixel $k \in \{1, \dots, n\}$, the per-branch losses are:
\begin{equation}
    \begin{aligned}
        L_{select} &= \tfrac{1}{n}\sum\nolimits_{k}(I_k^{HR} - I_k^{GT})^{\!2}, \qquad
        L_f = \tfrac{1}{n}\sum\nolimits_{k}(I_k^{HR_f} - I_k^{GT})^{\!2}, \\[3pt]
        L_i &= \tfrac{1}{n}\sum\nolimits_{k}\!\left(\tfrac{\mu}{\sigma}(I_k^{HR_i} - I_k^{GT})\right)^{\!2}, \quad i \in \{p,\, pf,\, fp\},
    \end{aligned}
\label{eq:loss_terms}
\end{equation}
where the $\mu / \sigma$ weighting on the perceptual-stream losses follows the one-step diffusion approximation of Kim~\etal~\cite{kim2024}, accounting for the signal-to-noise ratio of the diffusion process. The overall objective is the unweighted sum $L = L_{select} + L_f + L_p + L_{pf} + L_{fp}$. Training details including optimizer, learning rate schedule, and data augmentation are provided in the supplementary material. 

% === Student original section header (commented out by professor) ===
% \section{Experiment}
% \label{sec:experiment}
% In this section, we present the experimental results and discuss their implications.
% We first describe the experimental setup in \cref{subsec::setup}.
% Following that, we evaluate our FPLIA in the ASISR task across diverse datasets in \cref{subsec::validation}.
% Next, \cref{subsec::effectiveness} validate the effectiveness of combining fidelity-oriented and perceptual features by visualizing and analyzing the feature selections. Finally, \cref{subsec::complexity} presents the complexity analysis of FPLIA.
% === END student original section header ===

\section{Experiments}
\label{sec:experiment}
% This section evaluates FPLIA through quantitative benchmarks (\cref{subsec::quantitative}), qualitative comparisons (\cref{subsec::qualitative}), analysis of the internal feature selection behavior (\cref{subsec::selection_analysis}), and ablation studies (\cref{subsec::ablation}). \Cref{subsec::setup} first describes the experimental setup.

\subsection{Experimental Setup}
\label{subsec::setup}

The framework is trained on DIV2K~\cite{div2k} and Flickr2K~\cite{flickr2k}, which together provide 3,450 images at 2K resolution. 
Evaluation is conducted on the DIV2K validation set~\cite{div2k}, Set5~\cite{set5}, Set14~\cite{set14}, B100~\cite{b100}, and Urban100~\cite{urban100} across integer upscaling factors from $\times 4$ to $\times 16$; non-integer factors are additionally examined to validate continuous-scale generalization. 
% Among these benchmarks, Urban100 is particularly relevant to the structural fidelity challenge identified in \cref{sec:intro}, as its images are dominated by repetitive geometric patterns and fine architectural details that stress-test the fidelity-perception balance.  -> no image support this, maybe just remove
Reconstruction quality is measured by PSNR for pixel-level fidelity and LPIPS~\cite{lpips} for perceptual similarity, directly reflecting the two axes of the trade-off that motivates FPLIA. 
Throughout all tables, the best and second-best results are highlighted in \st{bold} and \nd{underline}, respectively.

FPLIA is compared against seven representative ASISR methods spanning both paradigms. The regression-based group includes LIIF~\cite{liif}, LTE~\cite{lte}, LIT~\cite{clit}, CiaoSR~\cite{ciaosr}, and HIIF~\cite{hiif}, all equipped with SwinIR~\cite{swinir} backbones, representing the state of the art in fidelity-oriented ASISR. The diffusion-based group includes IDM~\cite{idm} and Kim \etal~\cite{kim2024}; the latter is the most recent diffusion-based ASISR method and serves as the primary competitor. This selection covers the full fidelity-perception spectrum and enables direct assessment of how FPLIA bridges the two paradigms. For LIIF, LTE, CiaoSR, and HIIF, results are obtained using official pre-trained checkpoints. For LIT, IDM, and Kim \etal, models are retrained using official source code, as pre-trained checkpoints are not available.

\paragraph{Implementation details.}
The fidelity-oriented feature extractor $E_\theta$ is instantiated with SwinIR~\cite{swinir}, and the perceptual feature generator $G_\theta$ with Stable Diffusion v1.5~\cite{ldm}. The decoder $D_\theta$ is a five-layer MLP with GELU activation~\cite{gelu}. Both $E_\theta$ and $G_\theta$ are pretrained following their respective protocols~\cite{liif, ldm} and kept frozen during finetuning; only FPAM, FPSM, and $D_\theta$ are updated. The framework is backbone-agnostic, and results with alternative extractor and generator choices are reported in the supplementary material.
% [SUPP TODO] Supplementary should include: (1) optimizer (Adam) + lr schedule (5e-5, decay 0.5 at epochs 200/400/600/800) + batch size (32) + total epochs (1000); (2) patch cropping (48r x 48r HR / 48x48 LR) + augmentation (flip + 90 rotation); (3) pixel sampling (n=48^2) + five-branch loss formulas (Eq.9-10); (4) one-step diffusion approximation details + mu/sigma weighting; (5) alternative backbone experiments (different E_theta and G_theta combinations).

% === Student original §4.2 header (commented out by professor) ===
% \subsection{Validation of FPLIA}
% \label{subsec::validation}
% We validate the proposed framework FPLIA by comparing it with previous ASISR methods~\cite{liif,lte,clit,ciaosr,hiif,idm,kim2024}. For LIIF~\cite{liif}, LTE~\cite{lte}, CiaoSR~\cite{ciaosr}, and HIIF~\cite{hiif}, results are obtained using official pre-trained checkpoints from their respective GitHub repositories. For LIT~\cite{clit}, IDM~\cite{idm}, and Kim \etal~\cite{kim2024}, the results are generated by retraining the models using the official source code, as pre-trained checkpoints are not available.
% === END student original §4.2 header ===

\begin{figure*}[t]
    \centering
    \footnotesize
    % \begin{minipage}[t]{0.32\linewidth}
    %     \centering
    %     \centerline{\qquad \footnotesize DIV2K $\times$2}
    %     \vspace{-1pt}
    %     \includegraphics[width=\columnwidth]{figures/div2kx2.pdf}
    % \end{minipage}
    \begin{minipage}[t]{0.32\linewidth}
        \centering
        \centerline{\qquad \footnotesize DIV2K $\times$4}
        \vspace{-1.5pt}
        \includegraphics[width=\columnwidth]{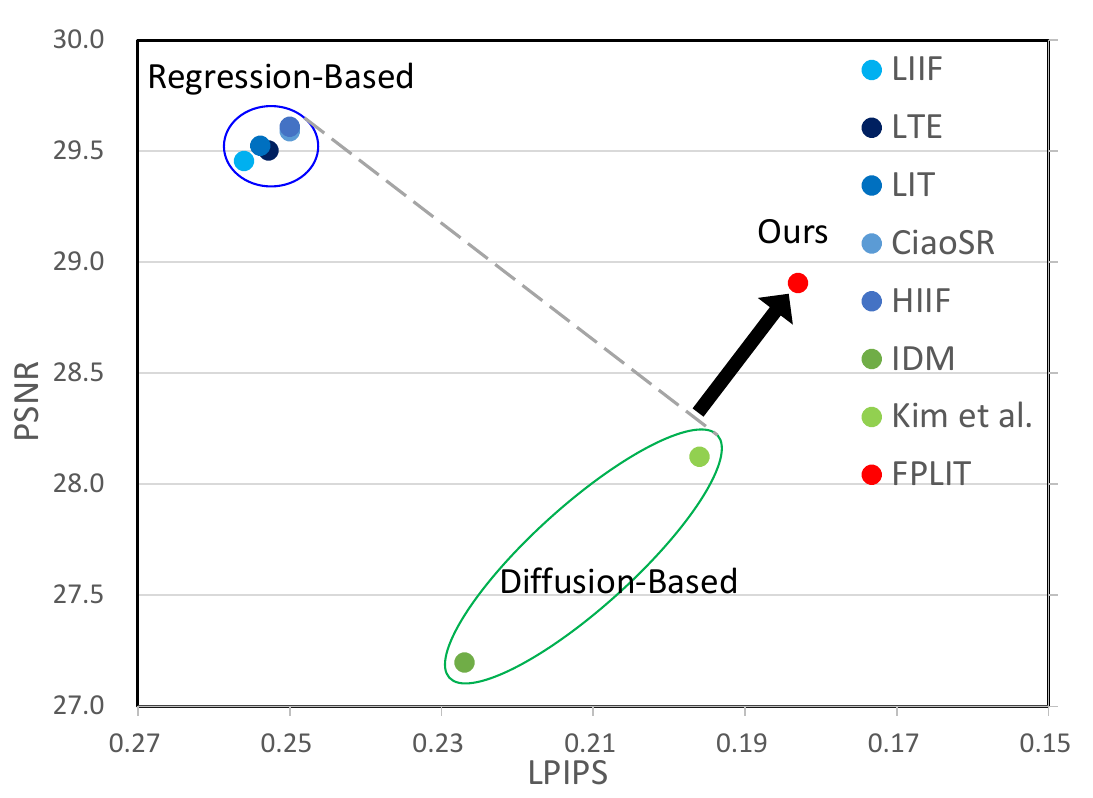}
    \end{minipage}
    \begin{minipage}[t]{0.32\linewidth}
        \centering
        \centerline{\qquad \footnotesize DIV2K $\times$6}
        \vspace{-1.5pt}
        \includegraphics[width=\columnwidth]{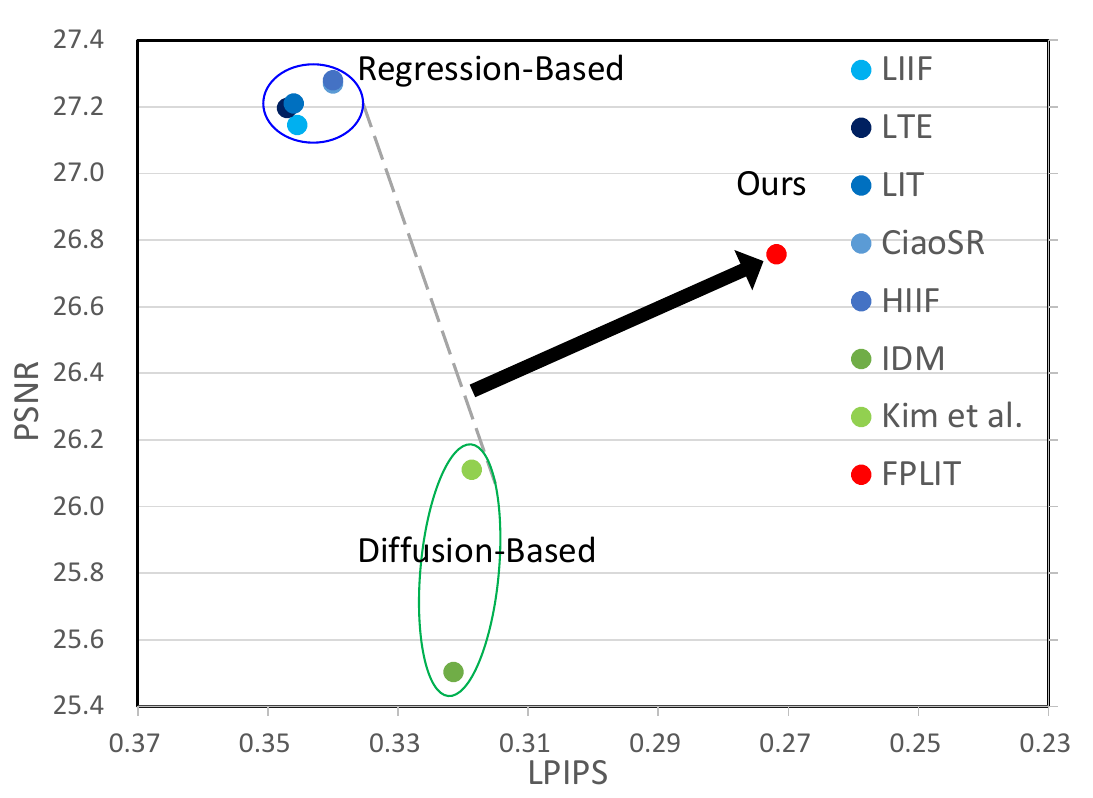}
    \end{minipage}
    \begin{minipage}[t]{0.32\linewidth}
        \centering
        \centerline{\qquad \footnotesize DIV2K $\times$8}
        \vspace{-1.5pt}
        \includegraphics[width=\columnwidth]{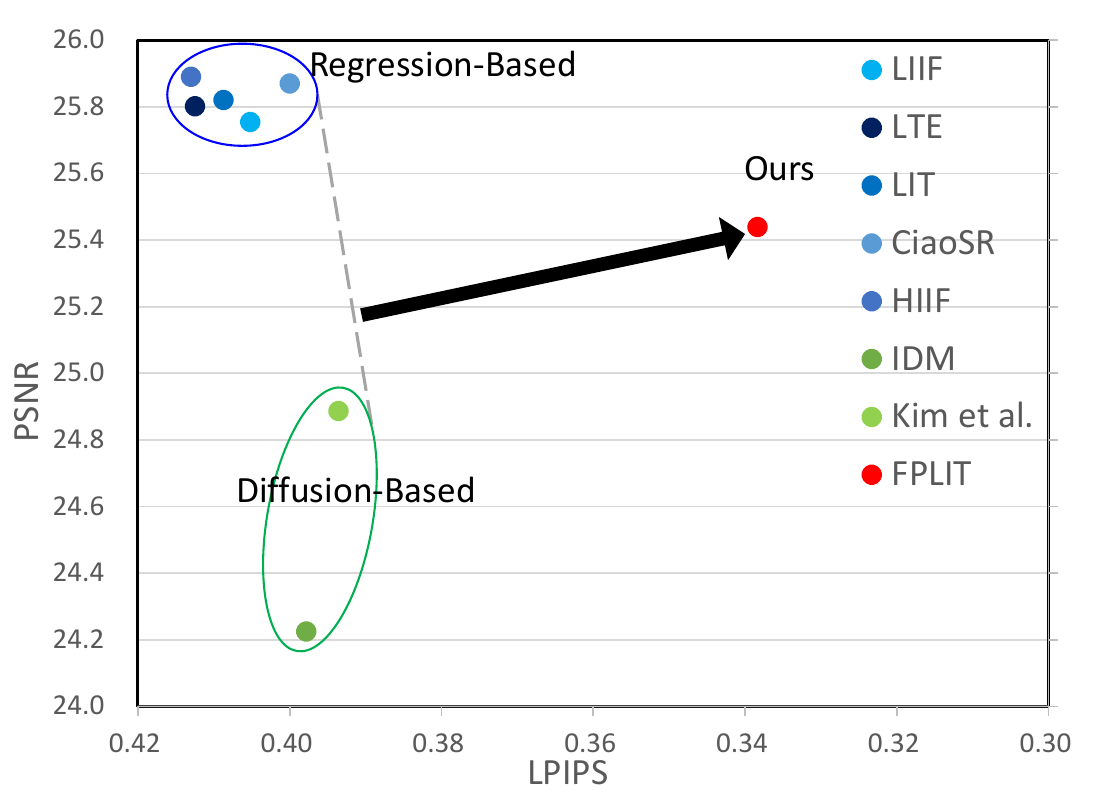}
    \end{minipage}
    % \\
    % \vspace{1pt}
    % \begin{minipage}[t]{0.32\linewidth}
    %     \centering
    %     \centerline{\qquad \footnotesize Set5 $\times$6}
    %     \vspace{-1.5pt}
    %     \includegraphics[width=\columnwidth]{figures/set5x6.pdf}
    % \end{minipage}
    % \begin{minipage}[t]{0.32\linewidth}
    %     \centering
    %     \centerline{\qquad \footnotesize Set5 $\times$8}
    %     \vspace{-1.5pt}
    %     \includegraphics[width=\columnwidth]{figures/set5x8.pdf}
    % \end{minipage}
    % \begin{minipage}[t]{0.32\linewidth}
    %     \centering
    %     \centerline{\qquad \footnotesize Set5 $\times$12}
    %     \vspace{-1.5pt}
    %     \includegraphics[width=\columnwidth]{figures/set5x12.pdf}
    % \end{minipage}
    \vspace{-2pt}
    \caption{
        Fidelity-oriented metric PSNR versus perceptual metric LPIPS across ASISR methods on DIV2K validation set~\cite{div2k} and Set5~\cite{set5} at wide-range upscaling factors. Our method (\textcolor{red}{red circle}) achieve the \textbf{Pareto frontier} between PSNR and LPIPS compared to regression-based methods (\textcolor{blue}{blue circle}) and diffusion methods (\textcolor[RGB]{0, 176, 80}{green circle}).
    }
    \label{fig::main}
    \vspace{-10pt}
\end{figure*}

\begin{table}[t]
\centering
\footnotesize
\caption{Comparison of regression-based and diffusion-based approaches with FPLIA. Average PSNR and LPIPS on Set14~\cite{set14}, B100~\cite{b100}, and Urban100~\cite{urban100}.}
\vspace{-10pt}
\resizebox{0.85\linewidth}{!}{
\begin{tabular}{cc|cc|cc|cc|cc|cc}
\toprule[1pt]
\multicolumn{2}{c|}{\multirow{2}{*}{Set5}}       & \multicolumn{2}{c|}{$\times$4}       & \multicolumn{2}{c|}{$\times$6}       & \multicolumn{2}{c|}{$\times$8}       & \multicolumn{2}{c|}{$\times$12}      & \multicolumn{2}{c}{$\times$16}       \\
\multicolumn{2}{c|}{}                            & PSNR $\uparrow$ & LPIPS $\downarrow$ & PSNR $\uparrow$ & LPIPS $\downarrow$ & PSNR $\uparrow$ & LPIPS $\downarrow$ & PSNR $\uparrow$ & LPIPS $\downarrow$ & PSNR $\uparrow$ & LPIPS $\downarrow$ \\
\midrule
\multicolumn{1}{c|}{}           & LIIF-SwinIR~\cite{liif}       & 32.71           & 0.167              & 29.44           & 0.233              & 27.33           & 0.288              & 24.90           & 0.386              & 23.26           & 0.474              \\
\multicolumn{1}{c|}{Regression} & LTE-SwinrIR~\cite{lte}        & 32.79           & 0.169              & 29.51           & 0.236              & 27.32           & 0.297              & 25.00           & 0.413              & 23.29           & 0.505              \\
\multicolumn{1}{c|}{}           & LIT-SwinrIR~\cite{clit}       & 32.82           & 0.168              & 29.49           & 0.235              & 27.38           & 0.292              & 24.88           & 0.399              & 23.35           & 0.483              \\
\multicolumn{1}{c|}{Methods}    & CiaoSR-SwinrIR~\cite{ciaosr}  & 32.83           & 0.167              & 29.57           & \textbf{0.231}     & 27.41           & \textbf{0.280}     & 25.06           & \textbf{0.378}     & 23.43           & \textbf{0.462}     \\
\multicolumn{1}{c|}{}           & HIIF-SwinIR~\cite{hiif}       & \textbf{32.87}  & \textbf{0.166}     & \textbf{29.63}  & 0.235              & \textbf{27.48}  & 0.296              & \textbf{25.13}  & 0.412              & \textbf{23.47}  & 0.498              \\
\midrule
\multicolumn{1}{c|}{Diffusion}  & IDM~\cite{idm}                & 30.11           & \textbf{0.110}     & 26.86           & \textbf{0.170}     & 24.67           & \textbf{0.246}     & 20.01           & 0.396              & 18.77           & 0.506              \\
\multicolumn{1}{c|}{Methods}    & Kim \etal~\cite{kim2024}     & \textbf{31.07}  & 0.128              & \textbf{27.74}  & 0.196              & \textbf{25.81}  & 0.284              & \textbf{22.57}  & \textbf{0.362}     & \textbf{21.50}  & \textbf{0.434}     \\
\midrule
\multicolumn{2}{c|}{FPLIA}                       & \textbf{31.95}  & \textbf{0.116}     & \textbf{28.73}  & \textbf{0.171}     & \textbf{26.76}  & \textbf{0.238}     & \textbf{24.24}  & \textbf{0.322}     & \textbf{22.83}  & \textbf{0.404}     \\
\multicolumn{2}{c|}{vs. Best Regression Method}  & -2.80\%         & +30.12\%           & -3.04\%         & +25.97\%           & -2.62\%         & +15.00\%           & -3.54\%         & +14.81\%           & -2.73\%         & +12.55\%           \\
\multicolumn{2}{c|}{vs. Best Diffusion Method}   & +2.83\%         & -5.45\%            & +3.57\%         & -0.59\%            & +3.68\%         & +3.25\%            & +7.40\%         & +11.05\%           & +6.19\%         & +6.91\%            \\
\bottomrule[1pt]
\toprule[1pt]
\multicolumn{2}{c|}{\multirow{2}{*}{Set14}}      & \multicolumn{2}{c|}{$\times$4}       & \multicolumn{2}{c|}{$\times$6}       & \multicolumn{2}{c|}{$\times$8}       & \multicolumn{2}{c|}{$\times$12}      & \multicolumn{2}{c}{$\times$16}       \\
\multicolumn{2}{c|}{}                            & PSNR $\uparrow$ & LPIPS $\downarrow$ & PSNR $\uparrow$ & LPIPS $\downarrow$ & PSNR $\uparrow$ & LPIPS $\downarrow$ & PSNR $\uparrow$ & LPIPS $\downarrow$ & PSNR $\uparrow$ & LPIPS $\downarrow$ \\
\midrule
\multicolumn{1}{c|}{}           & LIIF-SwinIR~\cite{liif}       & 28.95           & 0.272              & 26.82           & 0.384              & 25.33           & 0.431              & 23.35           & 0.530              & 22.20           & 0.585              \\
\multicolumn{1}{c|}{Regression} & LTE-SwinrIR~\cite{lte}        & 29.04           & 0.269              & 26.87           & 0.376              & 25.41           & 0.440              & \textbf{23.45}  & 0.546              & \textbf{22.30}  & 0.607              \\
\multicolumn{1}{c|}{}           & LIT-SwinrIR~\cite{clit}       & 29.00           & 0.269              & 26.84           & 0.375              & 25.38           & 0.437              & 23.39           & 0.537              & 22.27           & 0.595              \\
\multicolumn{1}{c|}{Methods}    & CiaoSR-SwinrIR~\cite{ciaosr}  & 29.08           & 0.269              & 26.87           & \textbf{0.371}     & 25.43           & \textbf{0.430}     & 23.44           & \textbf{0.521}     & 22.28           & \textbf{0.576}     \\
\multicolumn{1}{c|}{}           & HIIF-SwinIR~\cite{hiif}       & \textbf{29.10}  & \textbf{0.267}     & \textbf{26.98}  & 0.374              & \textbf{25.49}  & 0.440              & 23.43           & 0.542              & 22.29           & 0.605              \\
\midrule
\multicolumn{1}{c|}{Diffusion}  & IDM~\cite{idm}                & 27.13           & 0.221              & 25.11           & \textbf{0.299}     & 23.60           & \textbf{0.370}     & 20.74           & 0.485              & 19.03           & 0.573              \\
\multicolumn{1}{c|}{Methods}    & Kim\etal~\cite{kim2024}       & \textbf{27.77}  & \textbf{0.201}     & \textbf{25.91}  & 0.325              & \textbf{24.43}  & 0.389              & \textbf{22.29}  & \textbf{0.460}     & \textbf{20.89}  & \textbf{0.531}     \\
\midrule
\multicolumn{2}{c|}{FPLIA}                       & \textbf{28.48}  & \textbf{0.188}     & \textbf{26.50}  & \textbf{0.283}     & \textbf{25.01}  & \textbf{0.350}     & \textbf{23.01}  & \textbf{0.439}     & \textbf{21.74}  & \textbf{0.493}     \\
\multicolumn{2}{c|}{vs. Best Regression Method}  & -2.13\%         & +29.59\%           & -1.78\%         & +23.72\%           & -1.88\%         & +18.60\%           & -1.88\%         & +15.74\%           & -2.51\%         & +14.41\%           \\
\multicolumn{2}{c|}{vs. Best Diffusion Method}   & +2.56\%         & +6.47\%            & +2.28\%         & +5.35\%            & +2.37\%         & +5.41\%            & +3.23\%         & +4.57\%            & +4.07\%         & +7.16\%            \\
\bottomrule[1pt]
\toprule[1pt]
\multicolumn{2}{c|}{\multirow{2}{*}{B100}}       & \multicolumn{2}{c|}{$\times$4}       & \multicolumn{2}{c|}{$\times$6}       & \multicolumn{2}{c|}{$\times$8}       & \multicolumn{2}{c|}{$\times$12}      & \multicolumn{2}{c}{$\times$16}       \\
\multicolumn{2}{c|}{}                            & PSNR $\uparrow$ & LPIPS $\downarrow$ & PSNR $\uparrow$ & LPIPS $\downarrow$ & PSNR $\uparrow$ & LPIPS $\downarrow$ & PSNR $\uparrow$ & LPIPS $\downarrow$ & PSNR $\uparrow$ & LPIPS $\downarrow$ \\
\midrule
\multicolumn{1}{c|}{}           & LIIF-SwinIR~\cite{liif}       & 27.82           & 0.360              & 26.05           & 0.466              & 25.00           & 0.534              & 23.63           & 0.627              & 22.73           & 0.683              \\
\multicolumn{1}{c|}{Regression} & LTE-SwinrIR~\cite{lte}        & 27.85           & 0.355              & 26.08           & 0.469              & 25.02           & 0.545              & 23.64           & 0.648              & 22.77           & 0.710              \\
\multicolumn{1}{c|}{}           & LIT-SwinrIR~\cite{clit}       & 27.87           & 0.358              & 26.09           & 0.469              & 25.04           & 0.540              & 23.65           & 0.635              & 22.77           & 0.695              \\
\multicolumn{1}{c|}{Methods}    & CiaoSR-SwinrIR~\cite{ciaosr}  & 27.90           & 0.355              & 26.13           & \textbf{0.461}     & 25.06           & \textbf{0.532}     & 23.68           & \textbf{0.623}     & 22.77           & \textbf{0.677}     \\
\multicolumn{1}{c|}{}           & HIIF-SwinIR~\cite{hiif}       & \textbf{27.92}  & \textbf{0.351}     & \textbf{26.15}  & 0.467              & \textbf{25.09}  & 0.543              & \textbf{23.69}  & 0.644              & \textbf{22.80}  & 0.701              \\
\midrule
\multicolumn{1}{c|}{Diffusion}  & IDM~\cite{idm}                & 26.36           & 0.279              & 24.70           & \textbf{0.352}     & 23.68           & \textbf{0.430}     & 21.85           & \textbf{0.530}     & 18.46           & 0.668              \\
\multicolumn{1}{c|}{Methods}    & Kim\etal~\cite{kim2024}       & \textbf{26.64}  & \textbf{0.262}     & \textbf{25.16}  & 0.397              & \textbf{24.18}  & 0.466              & \textbf{22.68}  & 0.549              & \textbf{21.22}  & \textbf{0.608}     \\
\midrule
\multicolumn{2}{c|}{FPLIA}                       & \textbf{27.33}  & \textbf{0.251}     & \textbf{25.71}  & \textbf{0.354}     & \textbf{24.70}  & \textbf{0.431}     & \textbf{23.30}  & \textbf{0.520}     & \textbf{22.23}  & \textbf{0.569}     \\
\multicolumn{2}{c|}{vs. Best Regression Method}  & -2.11\%         & +28.49\%           & -1.68\%         & +23.21\%           & -1.55\%         & +18.98\%           & -1.65\%         & +16.53\%           & -2.50\%         & +15.95\%           \\
\multicolumn{2}{c|}{vs. Best Diffusion Method}   & +2.59\%         & +4.20\%            & +2.19\%         & -0.57\%            & +2.15\%         & -0.23\%            & +2.73\%         & +1.89\%            & +4.76\%         & +6.41\%            \\
\bottomrule[1pt]
\toprule[1pt]
\multicolumn{2}{c|}{\multirow{2}{*}{Urban100}}   & \multicolumn{2}{c|}{$\times$4}       & \multicolumn{2}{c|}{$\times$6}       & \multicolumn{2}{c|}{$\times$8}       & \multicolumn{2}{c|}{$\times$12}      & \multicolumn{2}{c}{$\times$16}       \\
\multicolumn{2}{c|}{}                            & PSNR $\uparrow$ & LPIPS $\downarrow$ & PSNR $\uparrow$ & LPIPS $\downarrow$ & PSNR $\uparrow$ & LPIPS $\downarrow$ & PSNR $\uparrow$ & LPIPS $\downarrow$ & PSNR $\uparrow$ & LPIPS $\downarrow$ \\
\midrule
\multicolumn{1}{c|}{}           & LIIF-SwinIR~\cite{liif}       & 27.14           & 0.200              & 24.59           & 0.298              & 23.14           & 0.375              & 21.42           & 0.501              & 20.41           & 0.595              \\
\multicolumn{1}{c|}{Regression} & LTE-SwinrIR~\cite{lte}        & 27.23           & 0.194              & 24.66           & 0.296              & 23.18           & 0.381              & 21.46           & 0.520              & 20.41           & 0.620              \\
\multicolumn{1}{c|}{}           & LIT-SwinrIR~\cite{clit}       & 27.21           & 0.195              & 24.68           & 0.295              & 23.20           & 0.375              & 21.51           & 0.502              & 20.47           & 0.600              \\
\multicolumn{1}{c|}{Methods}    & CiaoSR-SwinrIR~\cite{ciaosr}  & 27.42           & \textbf{0.187}     & \textbf{24.86}  & \textbf{0.281}     & \textbf{23.34}  & \textbf{0.357}     & 21.61           & \textbf{0.477}     & 20.54           & \textbf{0.567}     \\
\multicolumn{1}{c|}{}           & HIIF-SwinIR~\cite{hiif}       & \textbf{27.44}  & 0.188              & 24.81           & 0.302              & 23.32           & 0.385              & \textbf{21.62}  & 0.516              & \textbf{20.56}  & 0.612              \\
\midrule
\multicolumn{1}{c|}{Diffusion}  & IDM~\cite{idm}                & 25.13           & 0.208              & 23.15           & 0.324              & 22.01           & 0.409              & 20.53           & 0.523              & 19.57           & 0.610              \\
\multicolumn{1}{c|}{Methods}    & Kim\etal~\cite{kim2024}       & \textbf{26.42}  & \textbf{0.170}     & \textbf{24.07}  & \textbf{0.284}     & \textbf{22.74}  & \textbf{0.356}     & \textbf{21.05}  & \textbf{0.466}     & \textbf{20.03}  & \textbf{0.549}     \\
\midrule
\multicolumn{2}{c|}{FPLIA}                       & \textbf{26.87}  & \textbf{0.160}     & \textbf{24.44}  & \textbf{0.260}     & \textbf{23.06}  & \textbf{0.338}     & \textbf{21.34}  & \textbf{0.453}     & \textbf{20.31}  & \textbf{0.534}     \\
\multicolumn{2}{c|}{vs. Best Regression Method}  & -2.08\%         & +14.44\%           & -1.69\%         & +7.47\%            & -1.20\%         & +5.32\%            & -1.30\%         & +5.03\%            & -1.22\%         & +5.82\%            \\
\multicolumn{2}{c|}{vs. Best Diffusion Method}   & +1.70\%         & +5.88\%            & +1.54\%         & +8.45\%            & +1.41\%         & +5.06\%            & +1.38\%         & +2.79\%            & +1.40\%         & +2.73\%            \\
\bottomrule[1pt]

\end{tabular}
}
\vspace{-15pt}
\label{table::main}
\end{table}

\vspace{-5pt}
\subsection{Quantitative Results}
\label{subsec::quantitative}

The ASISR landscape has been characterized by a persistent dichotomy: regression-based methods cluster in the high-PSNR, low-perceptual-quality region, while diffusion-based methods occupy the complementary quadrant. \Cref{fig::main} visualizes this trade-off on the DIV2K validation set and Set5 across upscaling factors from $\times 4$ to $\times 12$, plotting PSNR against LPIPS for all compared methods. FPLIA consistently occupies the Pareto frontier, achieving a fidelity-perception balance that no existing method reaches from either paradigm alone. 
% This positioning is a direct consequence of the two-stage design: FPAM generates four candidate features with distinct fidelity-perception profiles at every queried coordinate, and FPSM adaptively commits to the most reliable candidate on a per-pixel basis, so the framework avoids collapsing into a single global trade-off point. -> no domnstrate here but in ablation?

\Cref{table::main} reports detailed results on Set14~\cite{set14}, B100~\cite{b100}, and Urban100~\cite{urban100} across upscaling factors from $\times 4$ to $\times 16$. Two complementary findings emerge from this comparison. Against regression-based methods, FPLIA achieves consistently lower LPIPS with only a modest PSNR concession. Among the regression baselines, CiaoSR~\cite{ciaosr} and HIIF~\cite{hiif} alternate as the strongest competitors, with CiaoSR leading on LPIPS and HIIF on PSNR, yet neither excels on both metrics simultaneously because their architectures rely exclusively on pixel-regression features and lack the perceptual stream necessary for high-frequency synthesis. FPLIA addresses this gap through the cross-attention pathways in FPAM, which enrich fidelity-anchored representations with perceptual diversity. FPLIA improves LPIPS by $29.6\%$ over the best regression-based method on Set14 at $\times 4$, while the overall PSNR trade-off remains within $2.6\%$.

Against diffusion-based methods, the result is more definitive: FPLIA improves both PSNR and LPIPS simultaneously across the majority of benchmark-scale combinations. Kim \etal~\cite{kim2024}, the strongest diffusion baseline, obtains competitive LPIPS through generative texture synthesis, but this comes at the cost of structural fidelity, with PSNR consistently falling below the leading regression methods. The root cause is the uniform application of generative processing to all spatial locations, including regions where pixel-regression features already suffice. FPSM's content-aware selection avoids this failure mode by restricting perceptual candidates to regions where high-frequency synthesis is genuinely needed.

% The performance advantage also follows a scale-dependent pattern consistent with the analysis in \cref{sec:intro}. On Set14, the LPIPS gain over the best regression method reaches $+29.6\%$ at $\times 4$ and remains above $+14\%$ even at $\times 16$, while the PSNR trade-off stays below $2.6\%$ across all scales. This trend confirms that fidelity-only features become increasingly insufficient at higher upscaling factors where the low-resolution input retains less high-frequency information, and that FPSM's scale-aware confidence estimation appropriately shifts the balance toward perceptual candidates as the scale increases. The content-aware spatial selection described in the preceding paragraph and the scale-dependent shift described here operate along orthogonal axes; their combination is what enables FPLIA to maintain its advantage across the full range of conditions evaluated. -> scale shift maybe discuss in ablation 

\begin{figure*}[t]
    \centering
    \includegraphics[width=0.9\textwidth]{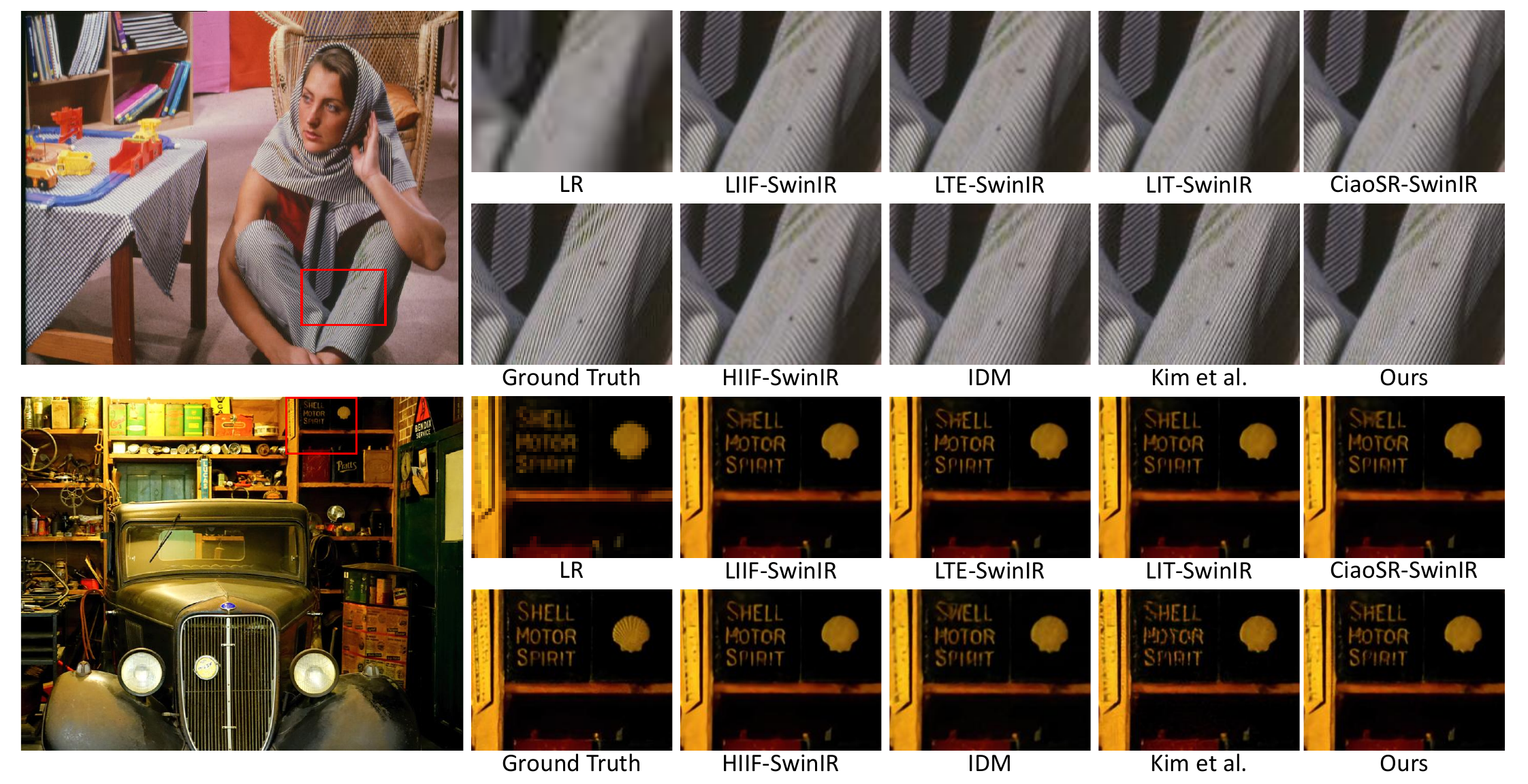}
    \caption{Comparison of the qualitative results of LIIF~\cite{liif}, LTE~\cite{lte}, LIT~\cite{clit}, CiaoSR~\cite{ciaosr}, HIIF~\cite{hiif}, IDM~\cite{idm}, Kim \etal~\cite{kim2024}, and our proposed FPLIA.}
    \label{fig::exp}
    \vspace{-5pt}
\end{figure*}

\begin{figure*}[t]
    \centering
    \includegraphics[width=0.9\textwidth,height=0.31\textheight]{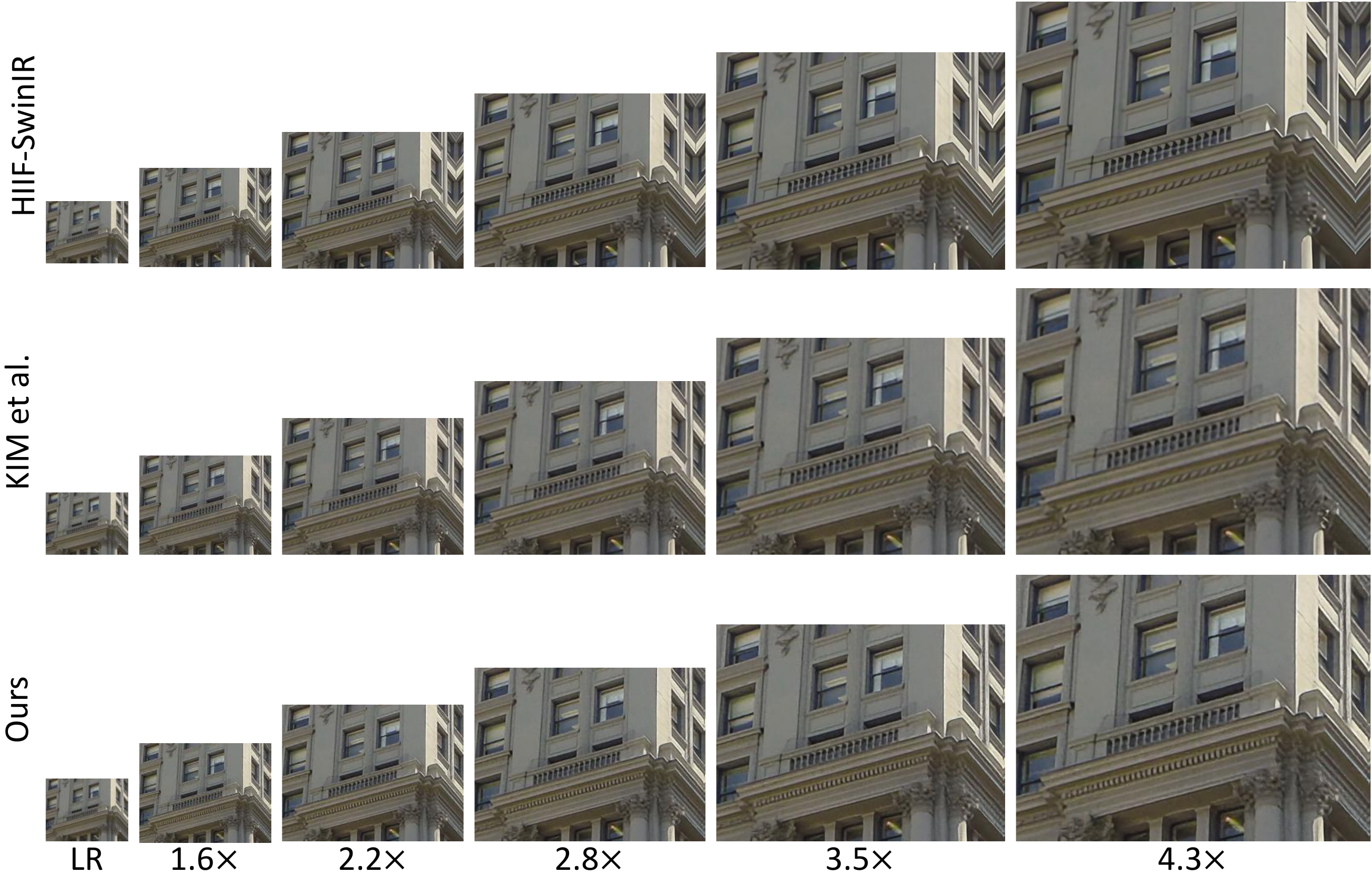}
    \caption{Comparison of the qualitative results of HIIF~\cite{hiif}, Kim \etal~\cite{kim2024}, and our proposed FPLIA with non-integer upscaling factors.}
    \label{fig::continuous}
    \vspace{-10pt}
\end{figure*}

% \vspace{-5pt}
\subsection{Qualitative Results}
\label{subsec::qualitative}

\Cref{fig::exp} presents visual comparisons against all seven baselines. Two capabilities of FPLIA stand out when examined alongside the failure modes of existing methods. For content with repetitive structural patterns, such as the striped texture in the first example, regression-based methods produce incorrect orientations and blurred reconstructions because their pixel-regression features lack the high-frequency diversity needed to resolve ambiguous periodic structures. The diffusion-based method of Kim \etal~\cite{kim2024} introduces noisy artifacts in the same region, as its generative process hallucinates textures without structural anchoring. FPLIA recovers the correct pattern with sharp detail, a result attributable to the bidirectional cross-attention in FPAM, which enables the fidelity stream to anchor the structural layout while the perceptual stream supplies high-frequency sharpness. For content requiring semantic consistency, such as the text in the second example, diffusion methods exhibit a different failure mode: IDM~\cite{idm} hallucinates spurious characters, and Kim \etal produce discontinuous letter forms. FPLIA avoids these artifacts through FPSM's content-aware selection, which favors fidelity-anchored candidates in semantically sensitive regions.

\Cref{fig::continuous} further validates the continuous-scale capability central to ASISR. On a building image upscaled through non-integer factors from $\times 1.6$ to $\times 4.3$, HIIF~\cite{hiif} and Kim \etal~\cite{kim2024} fail to faithfully reconstruct the architectural patterns, whereas FPLIA maintains consistent sharpness and structural correctness across the entire range. The coordinate-based local implicit formulation naturally accommodates arbitrary scales, and the adaptive selection performed by FPSM ensures that the fidelity-perception balance remains appropriate at each factor. 
\DeclareRobustCommand{\f}{\textcolor{blue}{\raisebox{-0.4ex}{\scalebox{2.0}{$\bullet$}}}}
\DeclareRobustCommand{\p}{\textcolor{green}{\raisebox{-0.4ex}{\scalebox{2.0}{$\bullet$}}}}
\DeclareRobustCommand{\pf}{\textcolor{red}{\raisebox{-0.4ex}{\scalebox{2.0}{$\bullet$}}}}
\DeclareRobustCommand{\fp}{\textcolor[RGB]{180, 20, 255}{\raisebox{-0.4ex}{\scalebox{2.0}{$\bullet$}}}}
\begin{figure}[t]
    \centering
    \includegraphics[width=0.9\columnwidth]{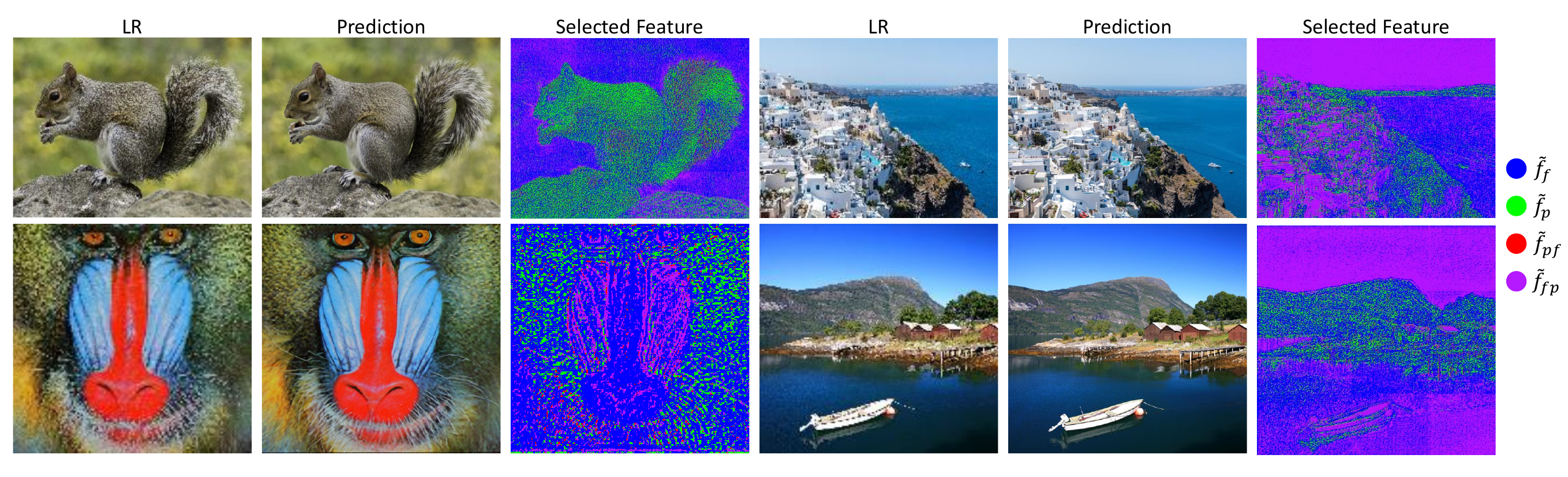}
    \caption{Visualization of the distribution of various selected features on a per-pixel basis, including fidelity feature $\tilde{f}_f$ (\f), perceptual feature $\tilde{f}_p$ (\p), perception-to-fidelity feature $\tilde{f}_{pf}$ (\pf), and fidelity-to-perception feature $\tilde{f}_{fp}$ (\fp).}
    \label{fig::select}
    % \vspace{-5pt}
\end{figure}

\begin{table}[t]
\centering
\footnotesize
\caption{Percentage of various selected features on DIV2K validation set~\cite{div2k}, including the fidelity feature $\tilde{f}_f$, perceptual feature $\tilde{f}_p$, perception-to-fidelity feature $\tilde{f}_{pf}$, and fidelity-to-perception feature $\tilde{f}_{fp}$.}
\vspace{-10pt}
\renewcommand{\arraystretch}{1.15}
\setlength{\tabcolsep}{5pt}
\resizebox{0.8\linewidth}{!}{
\begin{tabular}{c|cccccc}
\toprule[1pt]
Feature Branch      & $\times$2 & $\times$4 & $\times$6 & $\times$8 & $\times$12 & $\times$16 \\
\midrule
$\tilde{f}_f$       & 47.35\%  & 37.34\%  & 34.02\%  & 32.27\%  & 32.60\%   & 33.49\%   \\ 
$\tilde{f}_p$       & 28.92\%  & 19.09\%  & 16.96\%  & 16.97\%  & 17.05\%   & 17.53\%   \\
$\tilde{f}_{pf}$    & 1.21\%   & 2.42\%   & 2.91\%   & 3.29\%   & 3.64\%    & 3.87\%    \\
$\tilde{f}_{fp}$    & 22.52\%  & 41.15\%  & 46.11\%  & 47.48\%  & 46.71\%   & 45.11\%   \\
\bottomrule[1pt]
\end{tabular}
}
\label{table::select}
\vspace{-15pt}
\end{table}

\subsection{Analysis of Adaptive Feature Selection}
\label{subsec::selection_analysis}

This subsection examines the internal selection behavior of FPSM. \Cref{fig::select} visualizes the per-pixel feature selections on representative images. Along the spatial axis, FPSM consistently assigns the fidelity feature $\tilde{f}_f$ and the fidelity-to-perception feature $\tilde{f}_{fp}$ to smooth and background regions, while the perceptual feature $\tilde{f}_p$ is preferentially selected in areas with rich high-frequency textures. The selections remain spatially coherent across similar textures, avoiding the discontinuity artifacts that would arise from noisy or inconsistent confidence estimates.  

Along the scale axis, \cref{table::select} reveals a complementary trend in the global selection statistics. The proportion of coordinates assigned to $\tilde{f}_f$ decreases as the upscaling factor grows, while the proportion assigned to $\tilde{f}_{fp}$ increases. This shift demonstrates that as fidelity features become less informative at higher scales, FPSM increasingly relies on candidates that incorporate perceptual diversity. The perceptual feature $\tilde{f}_p$ follows a similar increasing trend, consistent with the greater reliance on generative modeling at large magnifications. The perception-to-fidelity feature $\tilde{f}_{pf}$ is selected at consistently low rates across all scales, aligning with the weaker independent performance of that branch observed in \cref{table::fpam}.

\subsection{Ablation Studies}
\label{subsec::ablation}

\begin{table}[t]
\centering
\footnotesize
\caption{Average PSNR and LPIPS on DIV2K validation set~\cite{div2k} of FPLIA with various component configurations.}
\vspace{-10pt}
\setlength{\tabcolsep}{5pt}
\resizebox{0.8\linewidth}{!}{
\begin{tabular}{cc|cc|cc|cc}
\toprule[1pt]
\multirow{2}{*}{FPAM} & \multirow{2}{*}{FPSM} & \multicolumn{2}{c|}{$\times$ 4}      & \multicolumn{2}{c|}{$\times$ 6}      & \multicolumn{2}{c}{$\times$ 8}       \\
                      &                       & PSNR $\uparrow$ & LPIPS $\downarrow$ & PSNR $\uparrow$ & LPIPS $\downarrow$ & PSNR $\uparrow$ & LPIPS $\downarrow$ \\
\midrule
-                     & -                     & 28.40           & 0.231              & 26.45           & 0.320              & 25.15           & 0.383              \\
$\checkmark$          & -                     & 28.53           & 0.188              & 26.55           & 0.304              & 25.28           & 0.371              \\
-                     & $\checkmark$          & 28.60           & 0.211              & 26.53           & 0.302              & 25.24           & 0.366              \\
$\checkmark$          & $\checkmark$          & \textbf{28.91}  & \textbf{0.183}     & \textbf{26.76}  & \textbf{0.272}     & \textbf{25.44}  & \textbf{0.338}     \\
\bottomrule[1pt]
\end{tabular}
}
\label{table::component}
\vspace{-5pt}
\end{table}

All ablation experiments are conducted on the DIV2K validation set~\cite{div2k} at upscaling factors $\times 4$, $\times 6$, and $\times 8$. \Cref{table::component} isolates the contributions of FPAM and FPSM by replacing each with its baseline counterpart: FPAM is substituted with the LIIF~\cite{liif} feature extraction pipeline, and FPSM with standard feature concatenation. Removing FPAM while retaining FPSM yields improved PSNR relative to the baseline without either module, yet LPIPS remains noticeably worse than the full model, confirming that the cross-feature interaction provided by FPAM is essential for generating candidates with sufficient perceptual diversity. Conversely, removing FPSM while retaining FPAM improves LPIPS over the no-module baseline but falls short of the full configuration, particularly at higher scales. This asymmetry is consistent with the design rationale in \cref{subsec:analysis}. The full model outperforms both partial configurations on every metric and scale, and the combined gain exceeds the sum of the individual improvements, indicating that the two modules operate synergistically rather than independently.

\begin{table}[t]
\centering
\footnotesize
\caption{Average PSNR and LPIPS on DIV2K validation set~\cite{div2k} of FPLIA using various combinations of the fidelity feature $\tilde{f}_f$, perceptual feature $\tilde{f}_p$, perception-to-fidelity feature $\tilde{f}_{pf}$, and fidelity-to-perception feature $\tilde{f}_{fp}$ in FPAM.}
\vspace{-10pt}
\renewcommand{\arraystretch}{1.15}
\setlength{\tabcolsep}{5pt}
\resizebox{0.8\linewidth}{!}{
\begin{tabular}{l|cc|cc|cc}
\toprule[1pt]
\multirow{2}{*}{Feature Branch}                                     & \multicolumn{2}{c|}{$\times$ 4}      & \multicolumn{2}{c|}{$\times$ 6}      & \multicolumn{2}{c}{$\times$ 8}       \\
                                                                    & PSNR $\uparrow$ & LPIPS $\downarrow$ & PSNR $\uparrow$ & LPIPS $\downarrow$ & PSNR $\uparrow$ & LPIPS $\downarrow$ \\
\midrule
$\tilde{f}_f$                                                       & \textbf{29.51}  & 0.258              & \textbf{27.22}  & 0.351              & \textbf{25.84}  & 0.414              \\
$\tilde{f}_p$                                                       & 28.12           & 0.194              & 26.11           & 0.310              & 24.89           & 0.380              \\
$\tilde{f}_{pf}$                                                    & 27.52           & 0.215              & 25.89           & 0.338              & 24.68           & 0.401              \\
$\tilde{f}_{fp}$                                                    & 28.28           & \underline{0.186}  & 26.31           &  \underline{0.289} & 25.03           & \underline{0.352}  \\
$\tilde{f}_f$ + $\tilde{f}_p$                                       & 28.71           & 0.206              & 26.61           & 0.299              & 25.31           & 0.361              \\
$\tilde{f}_f$ + $\tilde{f}_p$ + $\tilde{f}_{pf}$ + $\tilde{f}_{fp}$ & \underline{28.91}     & \textbf{0.183}     & \underline{26.76}     & \textbf{0.272}     & \underline{25.44}     & \textbf{0.338}     \\
\bottomrule[1pt]
\end{tabular}
}
\label{table::fpam}
\vspace{-15pt}
\end{table}

\Cref{table::fpam} further examines the role of each feature branch within FPAM. Using only the self-attention fidelity feature $\tilde{f}_f$ produces the highest PSNR across all scales but also the worst LPIPS, confirming that pixel-regression features alone carry no perceptual information. Adding the self-attention perceptual feature $\tilde{f}_p$ alongside $\tilde{f}_f$ yields only marginal LPIPS improvement, because the two self-attention outputs reside on separate manifolds and their naive combination lacks the cross-stream interaction needed to bridge the fidelity-perception gap. The decisive improvement arrives when the cross-attention features $\tilde{f}_{fp}$ and $\tilde{f}_{pf}$ are included: LPIPS drops markedly while PSNR remains competitive. This result validates the asymmetric interaction principle from \cref{sec:intro}: the directional information flow modeled by cross-attention produces candidates that neither self-attention branch can generate alone. 
\begin{table}[t]
\centering
\footnotesize
\caption{Average PSNR and LPIPS on DIV2K validation set~\cite{div2k} of FPLIA using various selection module in FPSM.}
\vspace{-10pt}
\setlength{\tabcolsep}{3pt}
\resizebox{0.8\linewidth}{!}{
\begin{tabular}{l|cc|cc|cc}
\toprule[1pt]
\multirow{2}{*}{Select Mechanism}    & \multicolumn{2}{c|}{$\times$ 4}      & \multicolumn{2}{c|}{$\times$ 6}      & \multicolumn{2}{c}{$\times$ 8}       \\
                                                           & PSNR $\uparrow$ & LPIPS $\downarrow$ & PSNR $\uparrow$ & LPIPS $\downarrow$ & PSNR $\uparrow$ & LPIPS $\downarrow$ \\
\midrule
Concatnation                                               & 28.53           & 0.188              & 26.55           & 0.304              & 25.28           & 0.371              \\
Spatial Feature Transform~\cite{sft}                       & 28.45           & \textbf{0.183}     & 26.51           & 0.307              & 25.23           & 0.375              \\
Per-pixel Selection                                        & \textbf{28.91}  & \textbf{0.183}     & \textbf{26.76}  & \textbf{0.272}     & \textbf{25.44}  & \textbf{0.338}     \\
Per-channel Selection                                      & 28.55           & 0.193              & 26.51           & 0.302              & 25.23           & 0.366              \\
\bottomrule[1pt]
\end{tabular}
}
\label{table::fpsm}
\vspace{-5pt}
\end{table}

\Cref{table::fpsm} compares four mechanisms for combining the candidate features before decoding: direct concatenation, Spatial Feature Transform (SFT)~\cite{sft} modulation, per-pixel selection, and per-channel selection. Concatenation performs worst overall, confirming the analysis in \cref{subsec:analysis} that linear fusion of features from different statistical manifolds is inherently suboptimal. SFT improves LPIPS at $\times 4$ by introducing spatially varying modulation weights, yet this soft blending still averages across candidates and fails to maintain the advantage at higher scales where the performance gap between candidates widens. Per-channel selection also underperforms per-pixel selection, because the fidelity-perception trade-off varies spatially rather than along the channel dimension. Per-pixel selection achieves the best results at every scale on both metrics, validating the discrete selection rationale developed in \cref{subsec:analysis}. 
\begin{table}[t]
\centering
\footnotesize
\caption{Inference latency and GPU memory footprint comparison.}
\vspace{-10pt}
\setlength{\tabcolsep}{3pt}
\resizebox{0.8\linewidth}{!}{
\begin{tabular}{l|cccc|c}
\toprule[1pt]
\multirow{2}{*}{Method}     & \multicolumn{4}{c|}{Latency (s)}                                           & GPU Memory   \\
                            & Fidelity Encoder & Perceptual Generator & Implicit Neural Function & Total & Footprint (GB) \\
\midrule
Kim \etal~\cite{kim2024}    & -                & 91.5 (LDM)           & 2.5 (IND)                & 94.0  & 25.1        \\
FPLIA                       & 2.2 (SwinIR)     & 92.1 (LDM)           & 4.1 (FPAM+FPSM+Decoder)  & 98.4  & 25.5        \\
\bottomrule[1pt]
\end{tabular}
}
\label{table::complexity}
\vspace{-15pt}
\end{table}

The performance gains from FPAM and FPSM come at minimal computational cost. \Cref{table::complexity} reports inference latency and GPU memory averaged over 100 runs of a $128 \times 128 \to 512 \times 512$ upscaling task on an NVIDIA A6000 GPU. Relative to the diffusion baseline of Kim \etal~\cite{kim2024}, FPLIA adds less than $5\%$ latency and less than $2\%$ peak memory. The overhead is small because FPAM operates on local coordinate neighborhoods rather than global feature maps, and FPSM consists of lightweight convolutional and fully connected layers. These figures confirm that the Pareto-frontier performance is achieved without meaningful efficiency penalty, closing the loop with the efficiency claim in \cref{sec:intro}.

\section{Conclusion}
\label{sec:conclusion}
This paper presents FPLIA, a novel framework that synergistically integrates fidelity-oriented features with a diffusion model to advance ASISR. FPLIA incorporates two architectural innovations including FPAM and FPSM to leverage both fidelity-oriented and perceptual features for realistic and faithful reconstructions. Extensive quantitative and qualitative evaluations across standard ASISR benchmarks with widely integer and non-integer upscaling factors substantiate the efficacy of these innovations. The results show that FPLIA achieves superior perceptual quality and lies on the Pareto frontier of perceptual quality versus reconstruction fidelity compared with state-of-the-art methods. Comprehensive ablation analyses further validate the contribution of each proposed component.


% \bibliographystyle{unsrt}
\bibliographystyle{splncs04}
\bibliography{references}

% \newpage
% \appendix
% \begin{center}
%     {\Large \textbf{Appendix}}
% \end{center}

% \input{sections/A0_supp}
% \input{sections/A1_related_work}
% \input{sections/A2_notation}
% \input{sections/A3_implement}
% \input{sections/A4_configure}
% \input{sections/A5_experiment}
% \input{sections/A6_futurework}

\end{document}